\documentclass{article}

\usepackage{PRIMEarxiv} 

\usepackage[dvipsnames]{xcolor}
\usepackage[utf8]{inputenc}
\usepackage[T1]{fontenc}
\usepackage{url}
\usepackage{booktabs}
\usepackage{amsfonts}
\usepackage{nicefrac}
\usepackage{microtype}
\usepackage{lipsum}
\usepackage{graphicx}
\graphicspath{{media/}}

\usepackage[numbers]{natbib}
\usepackage{amsmath, amssymb}
\usepackage{tikz}
\usepackage[absolute,overlay]{textpos}
\usepackage{listings}
\usepackage{makecell}
\usepackage{rotating}
\usepackage{multicol}
\usepackage{tcolorbox}
\tcbuselibrary{listings,breakable}
\usepackage{caption}
\usepackage{changepage}
\usepackage[misc,geometry]{ifsym}
\usepackage[toc,page]{appendix}

\definecolor{lightblu}{rgb}{0.9, 0.95, 1.0}

\usepackage{hyperref}

\title{Enhancing Surgical Documentation through Multimodal Visual-Temporal Transformers and Generative AI}
\author{
  Hugo Georgenthum\thanks{Hugo conducted all the experiments, conceptualization, tests, and code implementation. Cristian contributed to the conceptualization, theoretical analysis, and manuscript writing.} \\
  Department of Computer Science \\
  ETH Zürich \\
  Department of Computer Science \\
  Cambridge University \\
  \texttt{hugo.georgenthum@outlook.com}
  \And
  Cristian Cosentino\footnotemark[1] \\ 
  DIMES \\
  University of Calabria\\
  Department of Computer Science \\
  Cambridge University\\
  \texttt{ccosentino@unicalit}\\
  \texttt{cc2308@cam.ac.uk}
   \AND
  Fabrizio Marozzo \\
  DIMES\\
  University of Calabria\\
  \texttt{fmarozzo@dimes.unical.it}
   \And
  Pietro Liò \\
  Department of Computer Science \\
  Cambridge University \\
  \texttt{pietro.lio@cl.cam.ac.uk}
}

\begin{document}
\maketitle

\vspace{1cm}

\begin{abstract}
The automatic summarization of surgical videos is essential for enhancing procedural documentation, supporting surgical training, and facilitating post-operative analysis. This paper presents a novel method at the intersection of artificial intelligence and medicine, aiming to develop machine learning models with direct real-world applications in surgical contexts. We propose a multi-modal framework that leverages recent advancements in computer vision and large language models to generate comprehensive video summaries.
The approach is structured in three key stages. First, surgical videos are divided into clips, and visual features are extracted at the frame level using visual transformers. This step focuses on detecting tools, tissues, organs, and surgical actions. Second, the extracted features are transformed into frame-level captions via large language models. These are then combined with temporal features, captured using a ViViT-based encoder, to produce clip-level summaries that reflect the broader context of each video segment. Finally, the clip-level descriptions are aggregated into a full surgical report using a dedicated LLM tailored for the summarization task.
We evaluate our method on the CholecT50 dataset, using instrument and action annotations from 50 laparoscopic videos. The results show strong performance, achieving 96\% precision in tool detection and a BERT score of 0.74 for temporal context summarization. This work contributes to the advancement of AI-assisted tools for surgical reporting, offering a step toward more intelligent and reliable clinical documentation.
\end{abstract}

\hspace{1cm}


\clearpage

\section{Introduction}
\label{sec:intro}

Artificial Intelligence (AI) is playing an increasingly important role in modern surgery by enhancing precision, improving operational efficiency, and supporting decision-making. AI-based systems assist surgeons by automating routine tasks, analyzing large volumes of surgical data, and providing real-time guidance~\citep{hashimoto2018ai}. These capabilities contribute to reducing human error and improving patient outcomes, making AI an integral component of the surgical landscape.

A critical enabler of these systems is computer vision, which allows machines to interpret and understand complex visual data~\citep{maier2017surgical}. Convolutional Neural Networks (CNNs) have been widely used for such tasks due to their strong performance in learning spatial features from medical images~\citep{litjens2017survey}. They have been applied effectively to image-based diagnostics and intraoperative tool tracking. However, CNNs are limited in their ability to capture long-range dependencies across frames, which is crucial for understanding dynamic surgical procedures~\citep{zia2018surgical}.

Recent progress in deep learning has addressed this limitation with the introduction of transformer architectures. Transformers use self-attention mechanisms that enable models to incorporate both local and global context, making them particularly effective for analyzing sequential data~\citep{vaswani2017attention}. Originally developed for Natural Language Processing (NLP), transformers are the foundation of powerful Large Language Models (LLMs) like GPT, which can generate and understand human-like text~\citep{brown2020language}. These models are also being successfully applied to visual tasks, enabling richer and more contextual representations of video data~\citep{dosovitskiy2020image}.

Despite these advances, deploying AI in real-world surgical settings presents significant challenges. Procedures demand high precision, and even small errors can have serious consequences. AI models must therefore be accurate, transparent, and adaptable to diverse surgical environments. They must also meet strict ethical standards by preserving patient privacy, operating on consented data, and minimizing bias. Above all, these systems should assist rather than replace surgeons, ensuring human oversight in critical decisions.

At the same time, these constraints create opportunities for the development of autonomous AI agents capable of managing complex surgical workflows. With a robust modular design and the integration of explainable AI (XAI) techniques, such agents could orchestrate tasks like object detection, temporal analysis, and report summarization. By providing interpretable reasoning for each action, they would allow clinicians to monitor, guide, and adjust the system’s behavior while retaining full control over strategic decisions. This approach offers a promising path toward AI systems that are not only efficient but also safe, trustworthy, and clinically viable~\citep{holzinger2017we}.

In this work, we propose a multi-modal pipeline that combines computer vision and natural language processing to automatically generate structured surgical reports from video data. Our method involves three main stages. First, visual features are extracted from individual video frames using transformer-based models to detect instruments, anatomical structures, and surgical actions. Second, both the visual content and the frame-level captions are integrated using a ViViT-based encoder to model temporal patterns and generate clip-level summaries. Third, a specialized LLM composes a coherent and informative surgical report by aggregating the clip descriptions.

This approach distinguishes itself from existing methods by tightly integrating visual and linguistic information within a modular, context-aware architecture. The system is evaluated on a benchmark dataset of annotated laparoscopic surgery videos, which includes detailed annotations for tool usage and surgical actions. Experimental results indicate that the proposed pipeline achieves strong performance in both tool detection and temporal summarization tasks, validating the effectiveness of the multi-stage reasoning strategy.

The remainder of the paper is organized as follows. Section~\ref{sec:related} reviews the related work in surgical video analysis and AI-based summarization. Section~\ref{sec:methodology} describes the theoretical foundation and practical implementation of our proposed methodology. Section~\ref{sec:experiments} presents the experimental setup and analyzes the results. Finally, Section~\ref{sec:conclusion} concludes the paper by summarizing the main contributions and discussing directions for future research.

\section{Related Work}
\label{sec:related}

Automatic analysis of surgical videos is a rapidly evolving research area with significant potential in medical training, surgical skill assessment, and the continuous improvement of clinical practice \cite{loukas2018video, kawka2022intraoperative}. The widespread availability of intraoperative recordings, combined with advancements in computer vision and machine learning, has enabled systems capable of extracting semantic information from complex visual content. However, challenges remain due to the dynamic and multimodal nature of surgical videos—such as procedural variability, instrument occlusions, visual artifacts, and the need for accurate spatial-temporal context understanding \cite{jin2020multi, dimick2015surgical}.

Traditional approaches, focused solely on static visual features, have often proven inadequate in capturing the complexity of surgical actions, limiting both the quality of automatic summarization and the clinical usefulness of generated reports \cite{mao2022online,feng2024videooriontokenizingobjectdynamics}. Multimodal strategies have emerged as a promising direction, integrating heterogeneous data sources (visual, semantic, temporal) to produce more coherent and interpretable representations \cite{grenda2016using, augestad2020video}.

Recent advances in computer vision have enabled refined exploration of visual content in surgical videos, highlighting spatial, temporal, and semantic aspects. Deep learning models have evolved from convolutional neural networks (CNNs) to Vision Transformers, which can better model long-range dependencies between objects and actions \cite{islam2021exploring, khurana2021video}. A core component of surgical video analysis is object detection—identifying and localizing relevant items such as instruments, organs, or anatomical landmarks. Algorithms like YOLO, Faster R-CNN, and transformer-based methods have shown strong performance in complex surgical environments, which are often affected by narrow spaces, occlusions, and visual noise \cite{fu2018image, wang2025endochat}.

Based on these visual cues, video captioning techniques aim to generate textual descriptions that document surgical activities, aiding both education and clinical reporting. Recent models employ attention mechanisms and semantic alignment between frames and descriptions, improving accuracy and narrative consistency \cite{chen2023surgical, alsharid2022gaze}. Dense video captioning further refines this task by generating detailed, multi-temporal descriptions that capture transitions and surgical gestures as they unfold \cite{zhang2020dense, qasim2025dense}. These methods enhance interpretability and help detect recurring patterns or anomalies in procedures.

Building on object detection and feature extraction, captioning models aim to produce semantically rich and context-aware descriptions. In surgery, where precise interpretation is critical, these systems support training, documentation, and the development of explainable AI. Frame-level captioning focuses on describing individual frames, often combining spatial visual features with language models to form grammatically sound sentences. While early models used encoder-decoder architectures with CNNs and RNNs, recent approaches leverage pretrained multimodal models to better align image and text \cite{kutuk2024generating, fu2018image}.

However, frame-based captioning lacks temporal awareness. To overcome this, clip-level techniques model temporal dependencies and summarize sequences spanning multiple frames. These methods divide videos into short clips (e.g., 16–32 frames) and apply temporal modeling mechanisms like attention or graph-based reasoning to track action progressions \cite{zhang2019dilated, kashid2024stvs}.

Recent multimodal captioning frameworks fuse visual and semantic knowledge to improve accuracy and interpretability. For example, Chen et al. \cite{chen2023surgical} introduced a mutual-modal alignment model that learns from both modalities, generating clinically meaningful descriptions. In surgery, this is crucial for capturing terminology and sequence logic. Despite improvements, captioning systems still face challenges in maintaining coherence and completeness. Frame-level models may be too generic, while clip-level models risk inconsistency or missing details. Selecting the right captioning granularity involves a trade-off between descriptive depth and narrative flow—further complicated by surgical scene complexity and technical terminology \cite{sharma2023evolution}.

More recent medical AI efforts focus on combining visual and linguistic models to generate textual explanations from surgical content. Multimodal models allow joint representation learning by fusing visual sequences with semantic data \cite{rambhatla2022dl4burn, jujjavarapu2023predicting, wang2023novel}. Cross-modal attention has proven particularly effective, as it enables selective interaction between visual and language domains. For instance, the CroMA architecture \cite{antonio2024croma}, applied to robotic surgery VQA tasks, demonstrates how attention mechanisms can strengthen visual-textual alignment and improve response quality.

Large Language Models (LLMs) further enhance these capabilities by improving the fluency, coherence, and domain relevance of generated content. Studies show their potential in summarization, domain adaptation, and clinical explainability \cite{tian2023role, wang2023chatcad, hu2024advancing, van2024large}. Tools like ChatGPT and GPT-4 are being used to turn visual model outputs into structured surgical narratives aligned with medical terminology \cite{sloan2024automated}.

To mitigate information loss typical of frame-based models, VideoOrion \cite{feng2024videooriontokenizingobjectdynamics} proposes a dual-branch architecture—combining object-level and scene-level processing—to retain essential details and ensure a holistic understanding of video content. An important application of this integration is EndoChat\cite{wang2025endochat}, a multimodal system tailored for endoscopic surgery. It merges visual encoders with LLMs to generate contextualized surgical reports, effectively managing real-world complexities like tool occlusions, lighting variability, and scene clutter. This illustrates the value of synergy between modalities in producing meaningful outputs for clinical use.

Automatic surgical report generation is one of the most promising applications of these systems. Earlier methods based on templates or rules lacked flexibility. Today’s end-to-end models integrate multiple subtasks—object detection, captioning, and clip analysis—to generate comprehensive, context-aware summaries \cite{xu2021learning, chingnabe2025vision, bai2024m3d}. LLMs play a pivotal role, enabling fluent narrative construction, contextual adaptation, and knowledge injection. Techniques like domain adaptation and calibration \cite{xu2021learning} further refine output quality, tailoring it to clinical standards.

\begin{figure}[!ht]
    \centering
    \includegraphics[width=0.9\textwidth]{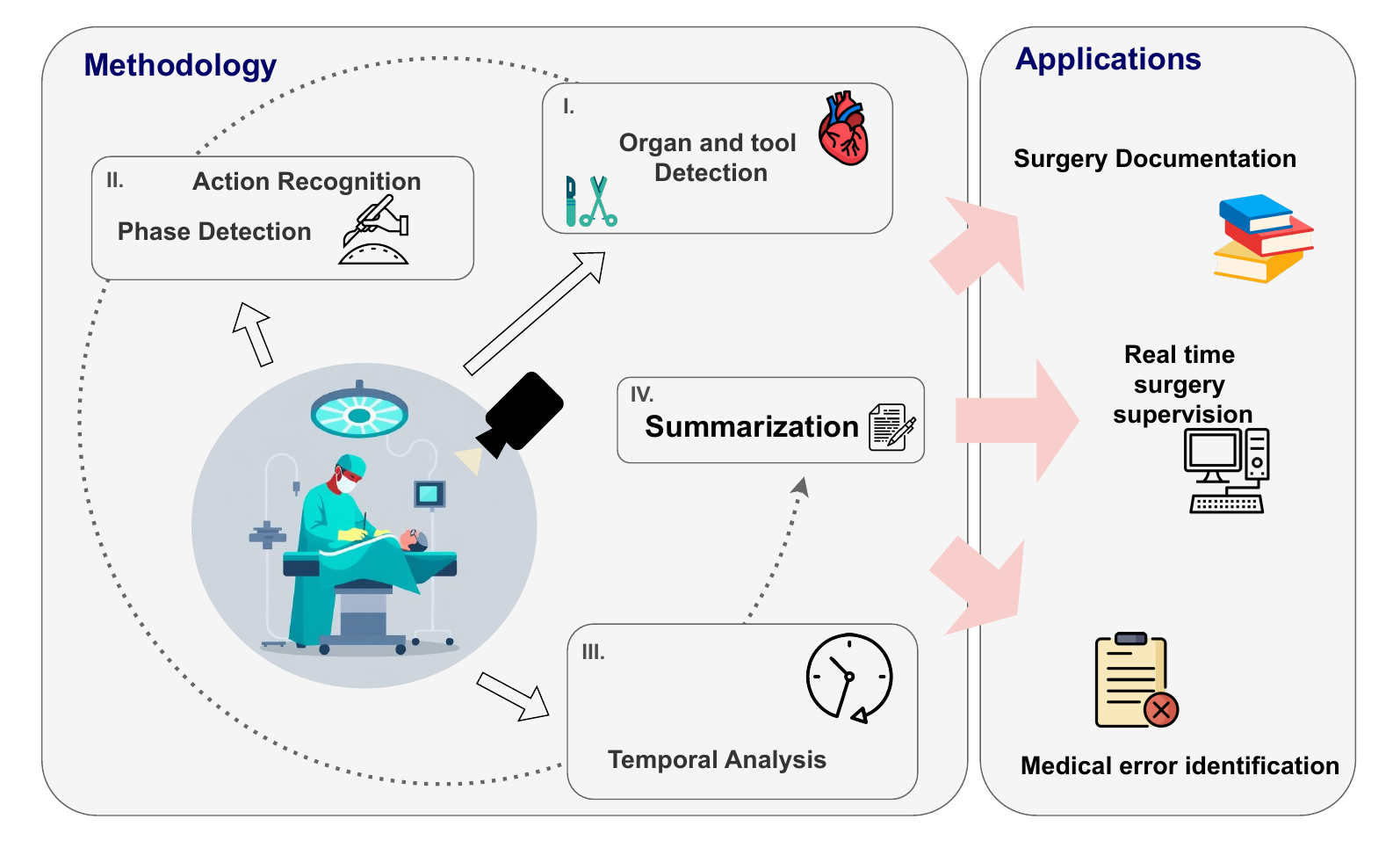}
     \caption{Overview of different modalities in surgical video and applications.}
    \label{fig:abstract}
\end{figure}

Despite the promising progress made in surgical video analysis—including multimodal captioning, object detection, and LLM-based summarization—current methods often treat these components in isolation or lack the architectural cohesion required for end-to-end report generation. Moreover, many models still overlook the need for interpretability across temporal scales and fail to tightly couple semantic scene understanding with clinical utility. Our work addresses these gaps by proposing an integrated pipeline (Figure~\ref{fig:abstract}) that spans from object and tool detection to action recognition, temporal modeling, and surgical summarization. Unlike prior approaches, we explicitly link frame-level and clip-level captioning through a multimodal fusion strategy, and we employ LLMs not just for summarization, but to translate procedural understanding into structured and explainable surgical reports. By doing so, we provide a unified framework that supports documentation, real-time surgical supervision, and medical error identification—bridging the divide between technical AI capabilities and real-world clinical application. The following section introduces our methodology in detail, highlighting how each component contributes to a comprehensive and explainable surgical reporting system.

\section{Proposed methodology}\label{sec:methodology}

This section presents our proposed methodology for the automated generation of explainable reports from surgical video analysis, an area of critical importance given the precision required in medical procedures, particularly in the surgical domain. Our goal is to accurately interpret surgical activities from videos, generate descriptive captions at multiple granularities, and produce comprehensive, human-readable summaries that clarify the model's reasoning and provide practical insights into surgical actions.

\begin{figure}[htb!]
    \centering
	\includegraphics[width=0.85\linewidth]{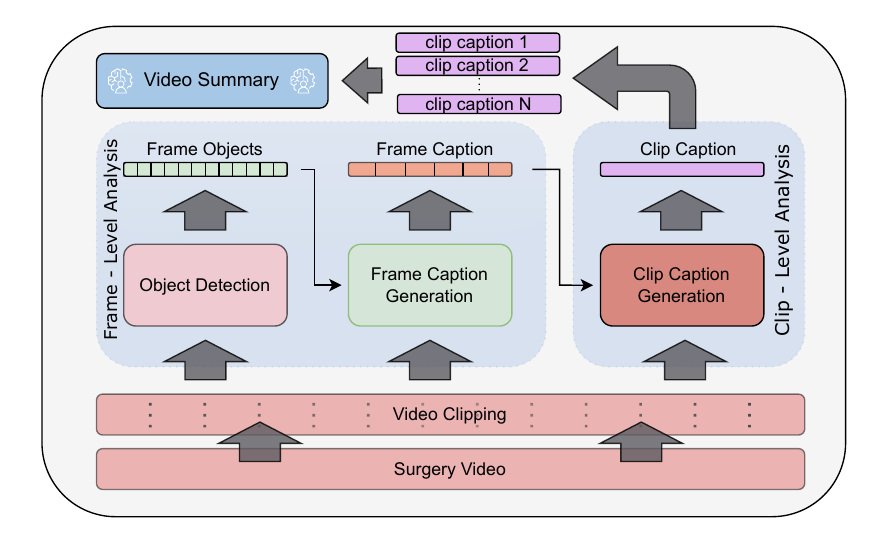}
	\caption{Execution flow of the proposed methodology.}
	\label{fig:proposed_framework}
\end{figure}

The methodology is structured into four key phases: (i) object detection in video frames; (ii) generation of descriptive captions at the frame level; (iii) creation of coherent captions at the clip level; and (iv) synthesis of a comprehensive surgical report. The execution flow of these steps is illustrated in Figure~\ref{fig:proposed_framework}.

The first phase involves detecting essential visual elements within each video frame, such as surgical instruments, organs, and anatomical landmarks. Given the complexity of surgical scenes—characterized by tight spaces, occlusions, and frequent motion blur—robust object detection models specifically adapted to the surgical domain are employed. Models like YOLO, Faster R-CNN, or customized transformer-based detectors are leveraged to accurately identify and localize key visual elements, providing the foundational context required for subsequent analysis.

The second phase focuses on generating precise captions at the frame level. Detected objects from the previous phase are combined with the original video frames, effectively forming a multimodal representation. This integration allows transformer-based captioning models to generate accurate and context-aware textual descriptions for individual frames. These captions explicitly identify surgical actions, objects involved, and spatial interactions within each frame, thereby capturing critical intra-frame information.

In the third phase, the methodology transitions from frame-level analysis to clip-level caption generation. Surgical videos, which can be extensive, are segmented into manageable temporal tokens—specifically clips containing 32 consecutive frames—to effectively model temporal dynamics. By treating frame captions generated in the previous phase as supplementary modalities, our model leverages temporal attention mechanisms to analyze inter-frame dependencies, capturing sequences of surgical actions that span across multiple frames. This results in the generation of coherent, temporally-aware captions that reflect complete surgical procedures or significant events within each clip.

The final phase synthesizes a human-readable surgical report, summarizing the entire surgical procedure in a clear, structured format. Utilizing generative AI, specifically large language models (LLMs) like GPT, individual clip captions are concatenated and processed into a comprehensive narrative summary. The LLM-based summarization explicitly leverages the context-rich clip captions, ensuring the resulting report is both coherent and interpretable. This approach bridges the gap between technical, granular AI-generated descriptions and intuitive, clinically relevant explanations that surgeons and medical professionals can readily utilize.

Through this integrated approach—encompassing detailed object detection, multimodal frame and clip captioning, and advanced generative summarization—our methodology not only provides precise automated analysis of surgical videos but also delivers meaningful, explainable reports. This enhances interpretability and usability in clinical settings, supporting informed decision-making and contributing to improved patient outcomes. In the following, we provide a detailed description of each of these methodological phases.

\subsection{Multi-Label Object Detection Using Vision Transformers}
\label{sec:vit-object-classification}

In this section, we detail the implementation of our multi-label object classification module using Vision Transformer (ViT) models, specifically adapted to surgical contexts~\cite{dosovitskiy2020image}. The primary objective is to detect multiple surgical objects such as instruments, organs, and anatomical structures—simultaneously within individual video frames. Initially, each input frame is transformed into standardized tensors of shape \( [3, 224, 224] \), corresponding to RGB color channels and image dimensions \((H, W)\). The Vision Transformer processes these tensors by dividing each image into a sequence of non-overlapping patches, each of dimension \(16 \times 16\). Consequently, for a standard input dimension, the total number of generated patches is calculated as:
\begin{equation}
N = \frac{H \times W}{p^2} = 196,
\end{equation}
where \( p=16 \) represents the patch size.

Given a matrix \( A = (a_{i,j}) \in \mathbb{R}^{n\times m} \), we define a submatrix or patch extraction as:
\[
A_{[i:i',j:j']} = \begin{bmatrix}
a_{i,j} & \dots & a_{i,j'} \\
\vdots & \ddots & \vdots \\
a_{i',j} & \dots & a_{i',j'}
\end{bmatrix}, \quad \text{for } 1\leq i\leq i' \leq n,\; 1\leq j\leq j' \leq m.
\]

Given an image matrix \( A\in \mathbb{R}^{n\times m} \), the \( k \)-th image patch \( P_k\in \mathbb{R}^{a\times b} \), assuming \( a,b \) evenly divide \( n,m \), respectively, is defined as:
\[
P_k = A_{[i:i+a-1,\,j:j+b-1]}, \quad \text{with} \quad k = i \times \frac{m}{b} + j.
\]
We denote the flattened representation of this patch as \( \tilde{P}_k \).

The input to the ViT, incorporating positional encoding, is formally represented as:
\begin{equation}
I = \big[ [1, \tilde{P}_1],\, [2, \tilde{P}_2], \dots, [N, \tilde{P}_{N}] \big] \in \mathbb{R}^{1 \times (16\times16+1)\times196}.
\end{equation}

Subsequently, the ViT generates an embedding processed through a softmax activation, producing a probability vector \( P \) for each class:
\begin{equation}
P = \text{softmax}(\text{ViT}(I)).
\end{equation}

The final set of detected objects is determined by applying a predefined threshold:
\begin{equation}
O = \{ \text{object}_i \mid P_i > \text{threshold} \}.
\end{equation}
Initially, a threshold of \( 0.5 \) is adopted, which can subsequently be calibrated based on empirical performance analysis~\cite{lin2017focal}.

Given the inherent class imbalance within surgical datasets—where certain objects appear infrequently—we implement a weighted Binary Cross-Entropy loss function to improve detection accuracy~\cite{cui2019class}. Formally, this loss is defined as:
\begin{equation}
\mathcal{L}\text{oss} = - \sum_{i=1}^{n} w_i \left[y_i \log \sigma(\hat{y_i}) + (1 - y_i) \log(1 - \sigma(\hat{y_i}))\right],
\end{equation}
where \( y_i \) and \( \hat{y_i} \) denote the ground truth and predicted logits, respectively. Class-specific weights \( w_i \) are computed as the inverse of class frequency \( f_i \):
\begin{equation}
w_i = \frac{1}{f_i + \epsilon},
\end{equation}
where \( \epsilon \) is a small constant introduced to prevent division by zero. Weights are normalized so their sum equals unity, ensuring balanced contributions during model optimization.

\subsection{Frame Caption Generation}
\label{sec:frame-caption-generation}

After detecting visual entities within individual video frames, our methodology proceeds with the extraction of semantic and contextual spatial information. Specifically, this phase aims to generate descriptive captions for each frame, emphasizing not only the detected surgical objects but also the actions performed by the surgeon within the surgical field. Transformer-based architectures are employed due to their proven capability to effectively capture intricate spatial and contextual dependencies in multimodal scenarios~\cite{vaswani2017attention, li2020oscar}. The architecture for frame caption generation comprises four main components:

\begin{itemize}
    \item \textbf{Frame Encoder:} Utilizes a pre-trained Vision Transformer (ViT) to encode each video frame into a rich latent representation. ViTs have demonstrated superior performance in capturing visual features by dividing images into fixed-size patches and modeling their interdependencies~\cite{dosovitskiy2020image, touvron2021training}. Given an input frame $I$, the ViT processes it into a sequence of embedded patches, subsequently projected into a 512-dimensional feature space via a learned linear transformation.

    \item \textbf{Object Encoder:} A textual encoder $E$, based on transformer architectures such as BERT or DistilBERT, is employed to process the set of detected object labels associated with each frame~\cite{devlin2018bert, Sanh2019DistilBERTAD}. This textual encoding ensures semantic information from detected objects is effectively integrated. The output embeddings from the textual encoder are similarly projected into a 512-dimensional latent space.

    \item \textbf{Feature Fusion:} To integrate visual and textual modalities, we fuse the encoded representations into a unified latent space. Fusion methods based on cross-modal attention have been successfully applied to enhance the joint representation of multimodal data~\cite{lu2019vilbert, li2020unicoder}.

    \item \textbf{Caption Decoder:} A transformer-based text decoder $D$ generates the final frame-level captions. Leveraging the cross-attention mechanism inherent to transformers, the decoder synthesizes meaningful and contextually precise descriptions based on integrated visual-textual representations~\cite{anderson2018bottom, cornia2020meshed}.
\end{itemize}

\begin{figure*}[!ht]
    \centering
	\includegraphics[width=0.8\textwidth, trim = 0 100 0 200, clip]{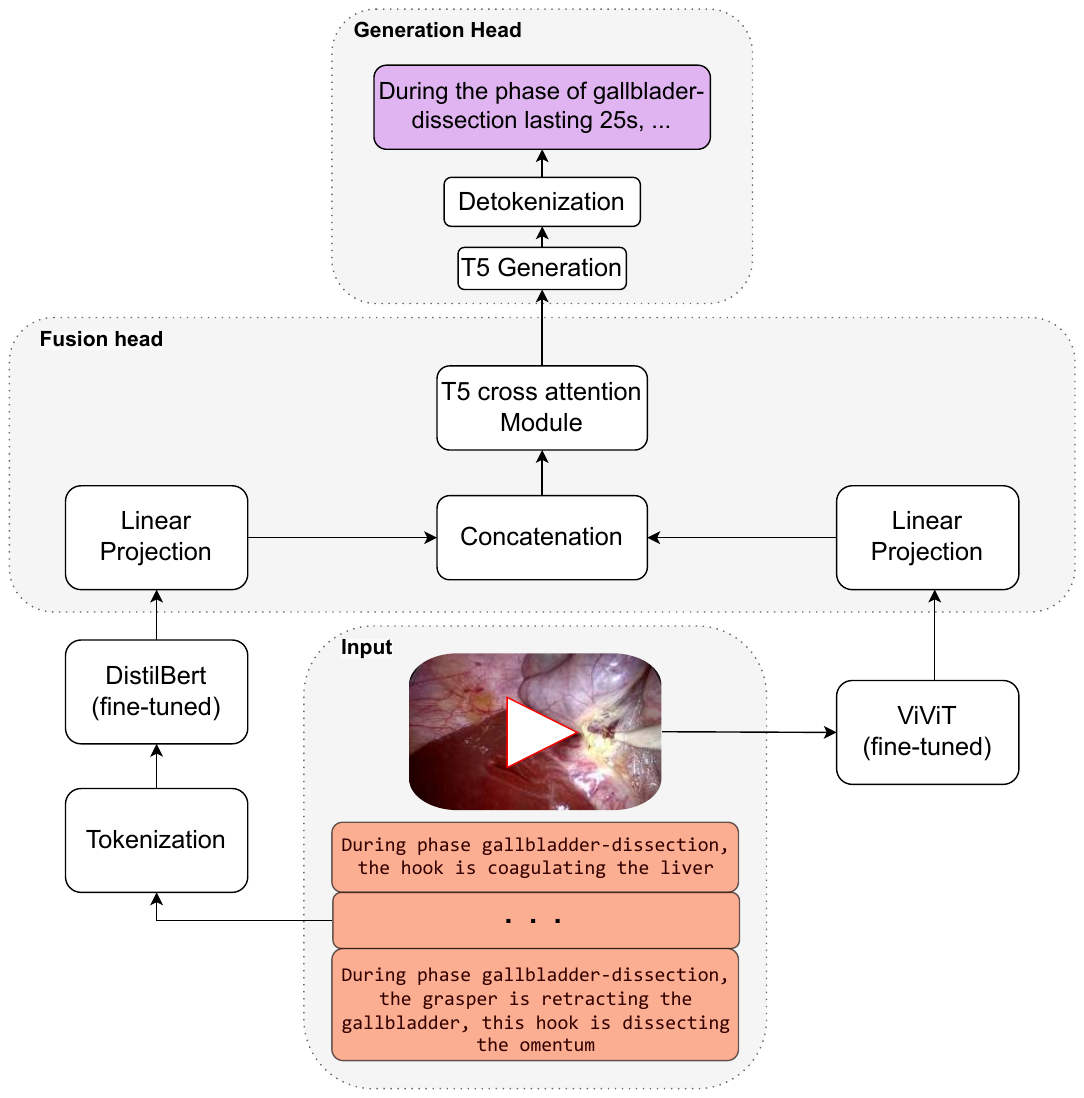}
	\caption{General scheme of the frame caption generation module.}
    \label{fig:generation_scheme}
\end{figure*}

Formally, given an image frame $I$ and the associated set of object labels $O$, the caption generation process is executed as follows:

\begin{align}
    h_I &= \text{ViT}(I) \quad \in \mathbb{R}^{N \times 768} \\
    h_I' &= W_I h_I \quad \in \mathbb{R}^{N \times 512} \\
    h_O &= E(O) \quad \in \mathbb{R}^{S \times 768} \\
    h_O' &= W_O h_O \quad \in \mathbb{R}^{S \times 512} \\
    h_C &= \text{concat}(h_I', h_O') \quad \in \mathbb{R}^{(N+S) \times 512} \\
    h_C' &= \text{CrossAttention}(h_C) \\
    C &= D(h_C')
\end{align}

Here, $W_I$ and $W_O$ represent learnable linear projection matrices for visual and textual features, respectively. The parameter $N$ denotes the total number of image patches extracted by the ViT encoder (consistent with the object detection phase described in Section \ref{sec:vit-object-classification}), while $S$ represents the number of tokens obtained from encoding object labels. The Figure~\ref{fig:generation_scheme} illustrates the general architecture of the proposed module for the automatic generation of captions, highlighting the effective integration of textual and visual representations through the combined use of DistilBERT and ViViT, with a cross-attention-based fusion module and a T5-based generative head.

This integrated approach ensures that the generated frame-level captions effectively encapsulate both visual and semantic contexts. By explicitly encoding spatial interactions and surgical actions within each frame, the methodology significantly enhances interpretability, thereby supporting clinical insights and facilitating informed decision-making in surgical procedures.

\subsection{Clip-Level Caption Generation}
\label{sec:clip-lvel-caption-generation}

Having effectively captured frame-level information, our methodology extends to modeling temporal dependencies across surgical video clips. Unlike frame-level analysis, clip-level captioning necessitates understanding actions and interactions that span multiple consecutive frames, thereby integrating temporal context into the generated descriptions. Inspired by recent advancements in video understanding literature, particularly transformer-based architectures such as Video Vision Transformers (ViViT)~\cite{arnab2021vivit}, our approach systematically addresses this requirement. Specifically, the clip-level caption generation phase differs from frame-level captioning primarily in two ways:

\begin{enumerate}
    \item \textbf{Temporal Integration}: While preserving the multimodal approach combining visual and textual inputs, the temporal dimension is explicitly introduced. Each input now comprises a sequence of frames organized into clips with dimensions \( [N_f, 3, H, W] \), where \( N_f \) represents the number of frames within each clip. Correspondingly, \( N_f \) frame captions previously generated serve as supplementary textual modalities.

    \item \textbf{Enhanced Patch Representation}: To effectively capture temporal dynamics, the vision transformer architecture extends spatial patches into spatiotemporal patches, integrating time as an additional dimension~\cite{bertasius2021space}. each patch is now represented by spatial dimensions augmented with temporal context, enabling the model to discern temporal variations across frames within a clip.
\end{enumerate}

Integrating these advancements allows our method to produce temporally coherent, contextually accurate clip-level captions. This ensures that surgical actions, particularly those evolving over multiple frames, are comprehensively described, significantly enhancing interpretability and clinical utility.

\subsection{Comprehensive Surgical Report Synthesis}
\label{sec:report-synthesis}

Following the generation of temporally coherent clip-level captions, the next phase involves synthesizing these captions into a comprehensive, structured surgical report. To achieve this, we employ large language models (LLMs), known for their ability to generate clear, coherent, and context-rich summaries~\cite{zhang2023benchmarking, radford2019language}. Specifically, we utilize GPT-based models, renowned for their proficiency in summarization and narrative construction~\cite{ouyang2022training}. The synthesis process involves feeding the entirety of generated clip captions into an LLM, accompanied by a carefully designed prompt explicitly instructing the model to produce a structured surgical summary. The chosen prompt guides the LLM to not only condense critical information but also to organize it in a clinically meaningful manner, suitable for direct utilization by medical professionals~\cite{brown2020language}.

Detailed considerations regarding prompt design and model selection are elaborated in Section~\ref{sec:Pre-trained Models}, highlighting the methodological rigor underlying our summarization approach. By employing this strategy, we ensure the resultant surgical reports effectively bridge technical AI-generated insights and human-readable, clinically actionable summaries. Thus, our comprehensive approach significantly enhances transparency, interpretability, and usability in surgical video analysis, ultimately facilitating improved clinical decision-making and patient care outcomes.

\section{Experimental Results}
\label{sec:experiments}

This section details the experimental pipeline of our surgical captioning system, structured into several consecutive stages. We begin by presenting the CholecT50 dataset and the preprocessing procedure used to convert surgical videos into structured representations at both the frame and clip levels (see Section~\ref{sec:dataset-preprocessing}). Next, we describe the experimental setup and introduce the pre-trained models employed for visual and textual feature extraction, highlighting their relevance in the context of minimally invasive surgery (see Section~\ref{sec:setting}). This is followed by a discussion of the training strategy used to optimize each module and mitigate error propagation.

To evaluate the system, we define a comprehensive set of metrics that assess both object classification accuracy and text generation quality (see Sections~\ref{sec:object-detection}, \ref{frame-caption}, and \ref{sec:clip-caption}). The results are presented across three core tasks: object detection, frame-level caption generation, and clip-level caption generation, each supported by both qualitative examples and quantitative analyses. We conclude by evaluating the final stage of our pipeline—structured surgical report generation—and discussing its completeness and consistency (see Section~\ref{sec:ev-report-generation}).

The GitHub repository containing the preprocessing scripts and trained models is publicly available\footnote{\url{https://github.com/Hugoggt/SurgeryVideoReportGeneration}}. The dataset used is licensed under the CC BY-NC-SA 4.0 license~\cite{cc-by-nc-sa-4.0}.

\begin{figure*}[!ht]
    \centering
	\includegraphics[width=1.1\textwidth, trim = 30 470 0 40, clip]{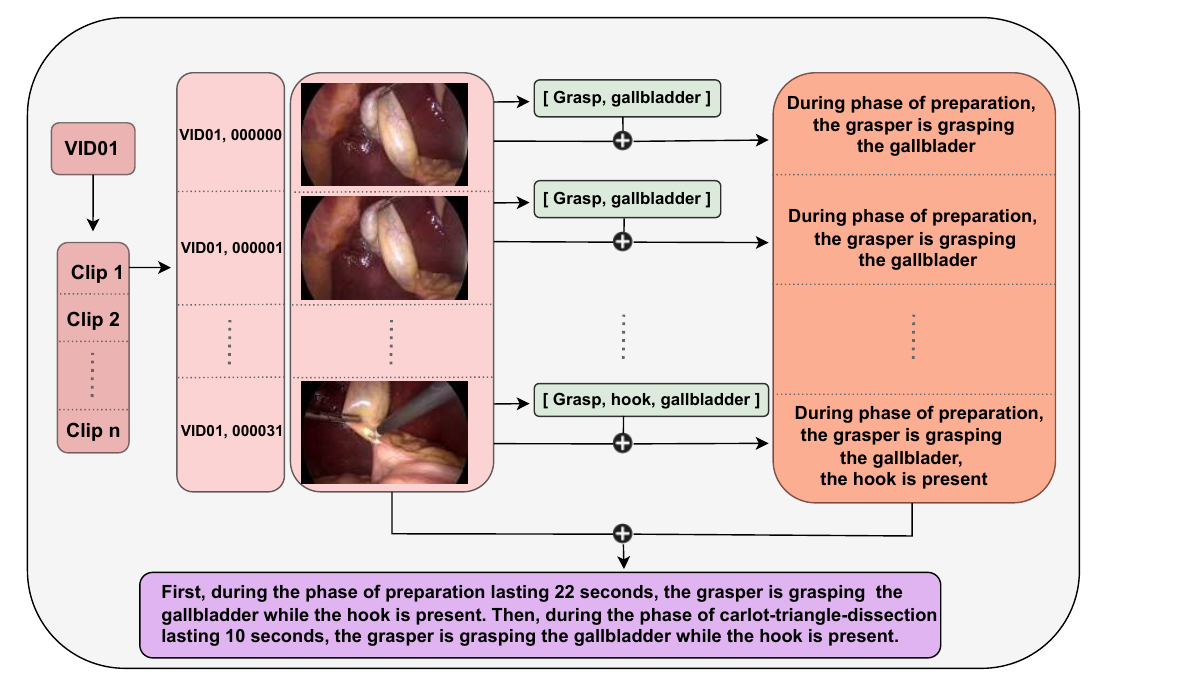}
	\caption{Visual walkthrough of the proposed workflow applied to a sample surgical video clip.}
    \label{fig:workflow}
\end{figure*}

Figure~\ref{fig:workflow} provides a visual walkthrough of the workflow applied to a sample clip from the first video (\texttt{VID01}, frames 000000–000031), showing the first 32 seconds of a laparoscopic cholecystectomy. The clip begins with the Preparation phase, where the grasper is used to lift and position the gallbladder for better visibility. This step helps the surgeon clearly identify anatomical landmarks and prepare the area for dissection.
The workflow then transitions into the Calot triangle dissection phase, marked by the introduction of the hook instrument. The surgeon then carefully dissects the tissue around the Calot triangle—the area formed by the cystic duct, common hepatic duct, and liver margin. This step is crucial for exposing the cystic duct and artery safely, reducing the risk of injury. The system’s frame-level captions and tool detections reflect these changes in activity and instrument usage.

\subsection{Dataset and Preprocessing}
\label{sec:dataset-preprocessing}

The CholecT50~\cite{cholect50} dataset used for the project has been provided by the University of Strasbourg and is composed of laparoscopic videos. A laparoscopic surgery is a procedure using a laparoscope, i.e. a thin tube equipped with a camera and a light. This kind of surgery is a subgroup of endoscopic procedures, often used for diagnostic purposes. The surgery in question is a \textit{cholecystectomy}, aiming to remove the gallbladder, involving small incisions and the use of a camera. This operation typically occurs because of the presence of gallstones which may cause pain or infections. 

The dataset is provided as 50 video folders, each video having a rate of 1 frame per second (fps), as well as tool, action, target and phase label for every frame.  In total, the dataset encompasses 100 distinct triplet categories derived from 6 instruments, 10 verbs, and 15 targets, resulting in over 151,000 annotated triplet instances. With a total of $89827$, the 
the clips are created by grouping the frames by $32$, with an overlap of $16$ frames between every clip, resulting in 6232 clips.

Frame caption and clip captions are not provided in the original dataset, and are created artificially using the annotations. To create the frame captions, the verb, target and phase are used to construct a simple sentence presenting what the surgeon is doing in the frame. The same goes for the clip caption construction; the frame captions are concatenated into one clip captions, taking into account the time and order each action is taken during the clip. 

\begin{figure}[!htt]
    \centering
    \begin{minipage}[b]{0.42\textwidth}
        \centering
        \includegraphics[width=\textwidth, trim = 0 240 0 230, clip]{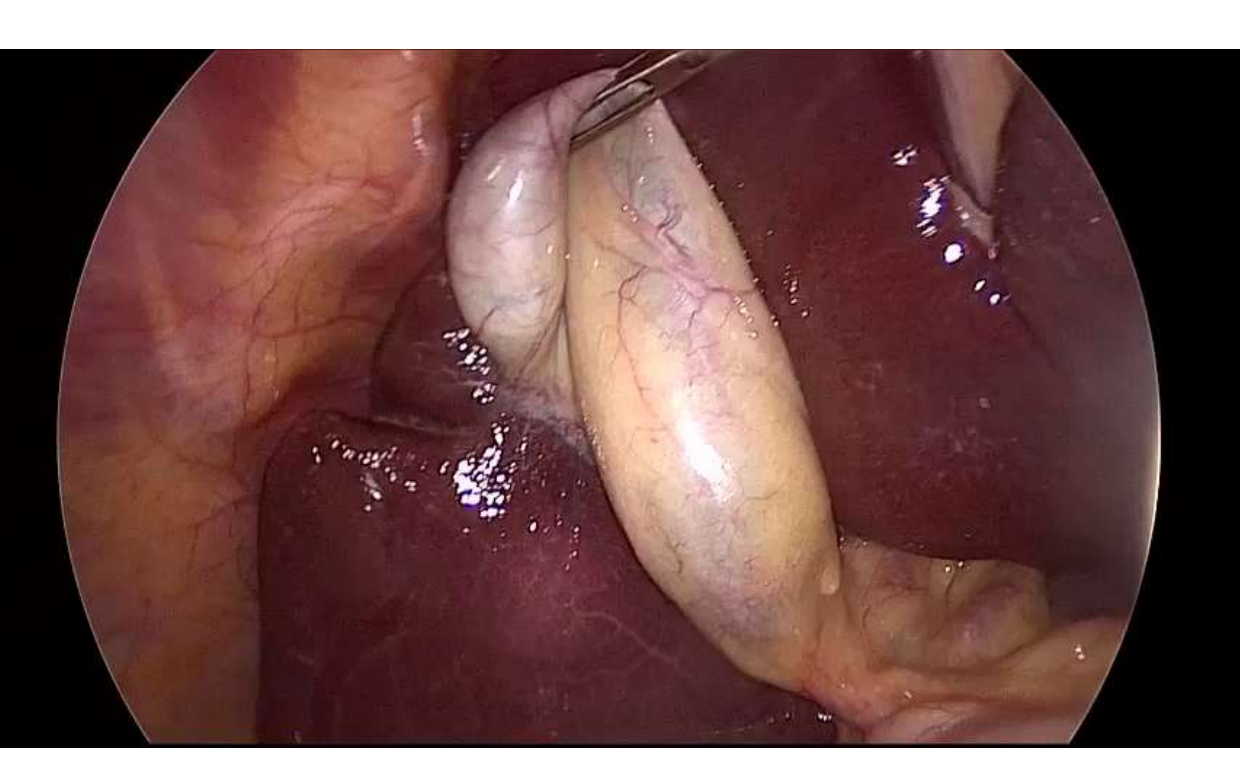}
    \end{minipage}
    \hfill
    \begin{minipage}[b]{0.43\textwidth}
        \begin{align*}
            &\textbf{Detected objects}: \text{grasper (forceps), gallbladder} \\
            &\textbf{Frame caption}: \text{During the preparation phase, the grasper} \\
            &\text{is holding the gallbladder.} \\
            &\textbf{Corresponding clip caption}: \text{First, during the 22-second} \\
            &\text{preparation phase, the grasper holds the gallbladder while} \\
            &\text{the hook is present. Then, during the 10-second} \\
            &\text{calot-triangle-dissection phase, the grasper continues} \\
            &\text{to hold the gallbladder while the hook remains present.}
        \end{align*}
        
    \end{minipage}
    \caption{Example of frame and clip caption generation from annotated surgical video.}
    \label{fig:frame-example}
\end{figure}

Figure \ref{fig:frame-example} illustrates how frame and clip captions are generated artificially from existing surgical annotations. On the left, a single frame is shown along with the recognized objects (for instance, the grasper and the gallbladder). On the right, a concise frame caption summarizes the immediate surgical action visible within that specific frame (e.g., the grasper holding the gallbladder), while a clip caption describes the broader context of the entire sequence (clip), including its duration, the surgical phase(s) in progress, and any other instruments present (such as the hook). This example demonstrates how individual frame-level descriptions can be combined to form a more comprehensive clip-level narrative, providing a detailed overview of both the action and the timeline associated with each phase of the procedure.

Having discussed how frame- and clip-level captions are generated to capture localized surgical actions, we now shift our focus to the overarching structure of the procedure. A $cholecystectomy$ can be divided into several distinct phases, each reflecting a specific surgical objective. Table \ref{tab:phase_counts} provides a detailed overview of these phases, listing the total number of frames that occur during each segment of the procedure (at a capture rate of 1 frame per second) as well as the corresponding duration in minutes. The phases include: (1) Preparation, (2) Calot-triangle-dissection, (3) Clipping-and-cutting, (4) Gallbladder-dissection, (5) Gallbladder-packaging, (6) Cleaning-and-coagulation, and (7) Gallbladder-extraction. The final row in the table summarizes the overall number of frames and cumulative time for the entire procedure.

\begin{table}[!ht]
\begin{adjustwidth}{-0.5in}{-0.5in} 
\centering
\fontsize{8pt}{9pt}\normalfont{
\begin{tabular}{@{\hspace{5pt}}l@{\hspace{5pt}}r@{\hspace{5pt}}r@{\hspace{5pt}}}
\toprule
\multicolumn{1}{c}{\textbf{Phase Name}} & \textbf{Frame Count} & \textbf{Time (minutes)} \\ 
\midrule
Preparation                & 2806   & 46.8   \\
Calot-triangle-dissection & 38808  & 646.8  \\
Clipping-and-cutting      & 7790   & 129.8  \\
Gallbladder-dissection    & 26789  & 446.5  \\
Gallbladder-packaging     & 3790   & 63.2   \\
Cleaning-and-coagulation  & 6986   & 116.4  \\
Gallbladder-extraction    & 2858   & 47.6   \\
\addlinespace[1ex]
\textbf{Total}            & \textbf{89927} & \textbf{1498.8} \\
\bottomrule
\end{tabular}
}
\caption{Overview of surgical phase durations and frame counts for a complete laparoscopic cholecystectomy. Time is reported in minutes based on a frame rate of 1 frame per second.}
\label{tab:phase_counts}
\end{adjustwidth}
\end{table}

The dataset is subsequently partitioned into three subsets: the \textit{training set}, comprising 80\% of the frames, and the \textit{test set} and \textit{validation set}, each accounting for 10\% of the frames.

\subsection{Setting up the experimental part}
\label{sec:setting}

This section outlines the experimental setup used to develop and evaluate our surgical captioning system. We begin by describing the pre-trained models adopted for visual and textual processing (see Section~\ref{sec:Pre-trained Models}), followed by the training strategy designed to improve robustness and mitigate error propagation (see Section~\ref{sec:training strategy}). Finally, we introduce the evaluation metrics employed to assess both classification and text generation performance (see Section~\ref{sec:evaluation metrics}).

\subsubsection{Pre-trained Models}
\label{sec:Pre-trained Models}

To effectively capture and summarize surgical video content, we leverage multiple pre-trained models, each selected for its specific strengths in processing visual and textual information. For encoding individual frames and video sequences, we employ two transformer-based vision architectures:

\begin{itemize}
    \item \textbf{ViT (Vision Transformer)}: We adopt the ViT-base-patch16-224-in21k model~\cite{dosovitskiy2021imageworth16x16words}, pre-trained on a large-scale image dataset. This model efficiently processes individual frames, extracting meaningful visual features crucial for frame-level understanding.
    \item \textbf{ViViT (Video Vision Transformer)}: To model temporal dependencies across video sequences, we employ the ViViT-B-16x2 model~\cite{arnab2021vivit}, pre-trained on the Kinetics-400 dataset. By extending ViT’s architecture to spatiotemporal domains, ViViT is particularly effective for analyzing complex procedural steps in surgical videos.
\end{itemize}

These models enable the effective extraction of both spatial and temporal features, resulting in a rich and comprehensive visual representation. For textual information regarding the detected surgical instruments and objects, we utilize \textbf{DistilBERT}~\cite{sanh2019distilbert}, a lightweight transformer model optimized for efficiency while preserving performance. As a distilled version of BERT, it offers a favorable balance between computational cost and encoding accuracy.

For the task of text generation, we integrate the T5 architecture~\cite{raffel2020t5}, well-suited for sequence-to-sequence applications. Specifically:

\begin{itemize}
    \item \textbf{T5-Small}: Employed to generate frame-level captions.
    \item \textbf{FLAN-T5-Base}: Used for producing coherent and context-aware captions at the surgical clip level.
\end{itemize}

The selection of T5 is motivated by its strong performance in text generation tasks, enabling the transformation of extracted features into fluent and informative surgical narratives. To further refine and structure the final surgical report, we evaluated various large language models (LLMs), including \textbf{Med-PaLM}~\cite{singhal2023expertlevelmedicalquestionanswering} developed by Google and \textbf{BioGPT}~\cite{10.1093/bib/bbac409} developed by Microsoft. While these models are fine-tuned for medical domains and demonstrate robust task-specific knowledge, their summarization capabilities are relatively limited.

As a result, the model selected for generating the final surgical report is \textbf{GPT-4}~\cite{openai2023gpt4,Cantini2025HarnessingPL,Cantini2025}, which exhibits excellent summarization capabilities alongside substantial domain-specific understanding. Notably, GPT-4 excels at producing outputs that adhere to structured reporting guidelines. Following several experimental evaluations, the prompt retained for guiding the generation process is defined as follows:

\begin{tcolorbox}[breakable=true,boxsep=0pt,left=2mm,right=1mm,top=1mm,bottom=1mm,
                  sharp corners, colback=lightblu, colframe=black, boxrule=0.5pt]

\textbf{Generate a concise and textual surgery report from the following sequential clip captions of a video.}  
Each clip describes a phase of the surgery, including the activity, tools used, and duration.

\textit{Key Instructions:}  

1. The clips form a continuous video. If multiple clips describe the same activity, combine their durations to reflect the total time spent on that activity.  

2. Write the report in a narrative format, explaining each phase step-by-step in a flowing text.

\textbf{Clip captions:} \{ clip captions \}
\end{tcolorbox}

By combining these models in a unified pipeline, we construct a robust framework for the automatic extraction, processing, and generation of informative and accurate descriptions of surgical procedures, thereby enhancing the quality and clarity of automated surgical reporting.

\subsubsection{Training Strategy}
\label{sec:training strategy}
The step by step clip caption generation $Objects \rightarrow Frame\ Caption \rightarrow Clip\ Caption$ benefits from the fact that each model is specialized in a specific task, using as input the output as the model before. However, models can still make mistakes. To improve the robustness of the models, we train the Frame Captioner ($FC$) to adapt to the Object Detector model ($OD$) errors. Similarly, the Clip Captioner ($CC$) is adapted to learn from FC's mistakes. In other words, considering input image $I$ and clip $Clip$ with the corresponding ground truth labels $O,\ F,\ C$, the training is divided in the two following parts. 

\begin{enumerate}
    \item \[ \tilde{O} = OD(I),\ \tilde{F} = FC(I,\ O),\ \tilde{C} = CC(Clip, F)
    \]
    In this part, each of the 3 models is trained independently from the others.
    \item \[ 
    \hat{F} = FC(I,\ \tilde{O}), \ \hat{C} = CC(clip, \hat{F})
    \]
    Now, instead of taking as input the ground truth objects $O$, FC takes as input the output of the object detection model. The same goes for the clip caption generation.
\end{enumerate}

The original $Cholec T50$~\cite{cholect50} is $59Gb$, and the processed dataset for frame- and clip-level analysis are respectively $50.44\ Gb$ and $111.86\ Go$. Due to the large size of the clip-level dataset, the training was done sequentially by separating the dataset in 2 and training one after an other using checkpoints.
The GPU used was $A100$, offering 83 Gb system RAM, 40 Gb GPU RAM and 235 Gb storage. The training of the heaviest model; the robust clip caption generator; lasted approximately 7 hours, with around 30 minutes per epoch. This slow training is due to the large amount of trainable parameters; the details for each model can be found here \ref{sec:modelparameters}.

\subsubsection{Evaluation Metrics}
\label{sec:evaluation metrics}

We evaluate the performance of our classification model using four standard metrics—precision, recall, F1 score, and accuracy. Precision measures how many of the predicted positives are actually correct, recall assesses how many of the actual positives are identified, the F1 score balances these two measures by combining them into a single indicator, and accuracy provides the overall proportion of correctly predicted instances.

To compare ourselves with other methods in the literature, we also introduce the average precision for instruments and targets, respectively noted as $AP_I$ and $AP_T$ and defined as
\[
\text{AP}_I = \frac{1}{C_I} \sum_{i=1}^{C_I} \text{AP}_i \ \ \ \ \ \ \ \ \ \ \ 
\text{AP}_T = \frac{1}{C_T} \sum_{j=1}^{C_T} \text{AP}_j
\]
with $C_I$ and $C_T$ the amount of instrument and target classes. For a class $i$ and given $R_n$ and $P_n$ recall and precision at threshold $n$: $AP_i = \sum_n (R_n - R_{n-1}) \cdot P_n$ being the area under the precision-recall curve.

Beyond these predictive performance metrics, we also examine how well the model’s predicted probabilities align with the actual outcomes by evaluating its calibration. Beyond measuring predictive performance, we also assess how well our model’s predicted probabilities align with actual outcomes (i.e., calibration). Expected Calibration Error (ECE) \cite{Guo2017} quantifies the difference between predicted confidence and the true likelihood of correctness. We partition the prediction confidences into \(M\) bins and compute:

\begin{equation}
\label{eq:ece}
\text{ECE} = \sum_{m=1}^{M} \frac{|B_m|}{n} \;\bigl|\;\text{acc}(B_m) - \text{conf}(B_m)\bigr|,
\end{equation}

where \(n\) is the total number of samples, \(|B_m|\) is the number of samples in bin \(m\), \(\text{acc}(B_m)\) is the average accuracy in bin \(m\), and \(\text{conf}(B_m)\) is the average predicted confidence in bin \(m\).

To mitigate overconfidence, \textit{temperature scaling} \cite{Guo2017} introduces a parameter \(T\) to rescale the model logits \(z\):

\begin{equation}
\tilde{z} = \frac{z}{T}.
\end{equation}

When \(T > 1\), the predicted probabilities are “smoothed.” The optimal \(T^*\) is typically found by minimizing the negative log-likelihood on a small validation set:

\begin{equation}
T^* = \arg\min_T \sum_{i} - y_i \log \bigl(\text{softmax}\bigl(\tfrac{z_i}{T}\bigr)\bigr).
\end{equation}

This approach recalibrates predictions while preserving classification accuracy. We also measure the similarity between candidate and reference texts using three widely adopted metrics: \textbf{BLEU} \cite{Papineni2002}, \textbf{ROUGE} \cite{Lin2004}, and \textbf{BERTScore} \cite{Zhang2019}.

\begin{itemize}
    \item \textbf{BLEU} calculates \(n\)-gram precision, penalizing excessively short outputs via a brevity penalty:
    \[
    \text{BLEU}_n = BP \cdot \exp \Bigl( \sum_{i=1}^{n} w_i \log p_i \Bigr),
    \]
    where \(p_i\) is the clipped precision for \(i\)-grams, \(w_i\) are weights, and 
    \[
    BP = \begin{cases} 
    1, & \text{if } c > r,\\ 
    e^{(1 - r/c)}, & \text{if } c \le r,
    \end{cases}
    \]
    with \(c\) being the length of the candidate text and \(r\) the length of the closest reference text.

    \item \textbf{ROUGE} evaluates how much of the reference text is captured by the candidate. We employ ROUGE-1 (unigram), ROUGE-2 (bigram), and ROUGE-L (longest common subsequence):
    \[
    \text{ROUGE-1} = \frac{\text{\# of overlapping unigrams}}{\text{\# of unigrams in reference}},
    \quad
    \\
    \text{ROUGE-2} = \frac{\text{\# of overlapping bigrams}}{\text{\# of bigrams in reference}},
    \quad 
    \\ 
    \]
    \[
    \text{ROUGE-L} = \frac{LCS(\text{candidate}, \text{reference})}{\text{length of reference}}.
    \]

    \item \textbf{BERTScore} compares contextual embeddings of tokens via cosine similarity. Let \(\text{predicted tokens} = P\) and \(\text{reference tokens} = R\). The BERTScore precision and recall are:
    \[
    \text{Precision} = \frac{1}{|P|} \sum_{t \in P} \max_{r \in R} \text{cosine\_similarity}(t, r),
    \quad
    \text{Recall} = \frac{1}{|R|} \sum_{r \in R} \max_{t \in P} \text{cosine\_similarity}(r, t),
    \]
    and the F1 score follows the standard harmonic mean formula.
\end{itemize}

These metrics capture various facets of textual alignment, from exact \(n\)-gram matching to deeper semantic similarity, thereby offering a comprehensive evaluation of the generated text.

\subsection{Object Detection}
\label{sec:object-detection}

The initial calibration performance of the model is already noteworthy, with an Expected Calibration Error (ECE) of $0.0075$. This low value indicates that the model's confidence estimates are generally well-aligned with the observed accuracy, suggesting that the model is not significantly overconfident or underconfident in its predictions. However, to further improve this alignment, temperature scaling—a post-hoc calibration technique—was applied. As a result, the ECE was reduced to $0.0028$, signifying a substantial improvement in calibration quality.

The temperature parameter that yielded the optimal calibration was $T = 1.8584$. This value effectively reduced overconfidence by smoothing the predicted probability distributions, particularly in higher-confidence bins. Such smoothing is essential in domains like surgical video analysis, where overconfident yet incorrect predictions can compromise interpretability and reliability.

\begin{figure}[!ht] \centering \includegraphics[width=0.8\textwidth]{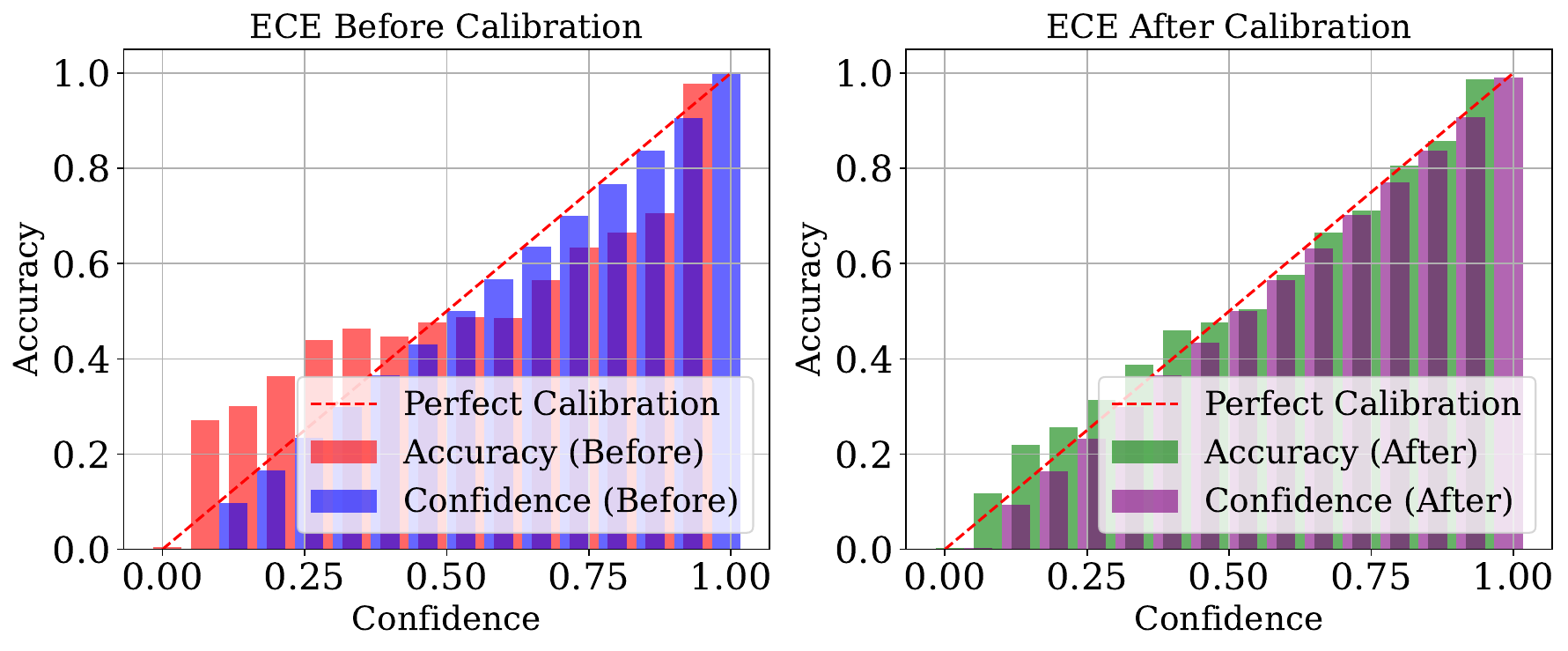} \caption{Reliability diagrams showing Expected Calibration Error before (left) and after (right) temperature scaling. The red dashed line represents perfect calibration. Bars indicate bin-wise accuracy and average confidence.} \label{fig:ece} \end{figure}

Figure~\ref{fig:ece} visually illustrates this calibration improvement using two reliability diagrams. The left panel (``ECE Before Calibration'') displays the bin-wise comparison between predicted confidence (in blue) and actual accuracy (in red) before applying temperature scaling. It can be observed that in some confidence intervals, particularly in the mid-to-high confidence range (e.g., $[0.6, 0.9]$), the confidence bars slightly exceed the corresponding accuracy bars, indicating overconfident predictions.

In contrast, the right panel (``ECE After Calibration'') demonstrates the post-calibration alignment between confidence (purple) and accuracy (green). The vertical bars are more consistently aligned and closely follow the ideal calibration line (dashed red), which represents perfect calibration (i.e., confidence equals accuracy). This visual confirmation supports the quantitative gain obtained through ECE reduction.

The calibration results were obtained by fixing a classification threshold of $0.5$, selected based on empirical evaluation to balance calibration performance with prediction precision. This threshold ensured consistent behavior across the confidence spectrum and allowed fair comparisons with prior works.

To further assess the performance of our approach, we compared it with two state-of-the-art models trained on the same dataset for action-verb-target triplet recognition in surgical videos. The first model, SurgT~\cite{surgt}, integrates a ResNet-18~\cite{he2016deep} backbone with a multi-task learning strategy, while the second model leverages a generative diffusion framework to generate surgical action triplets.

\begin{figure}[!h] \centering \includegraphics[width=0.4\textwidth]{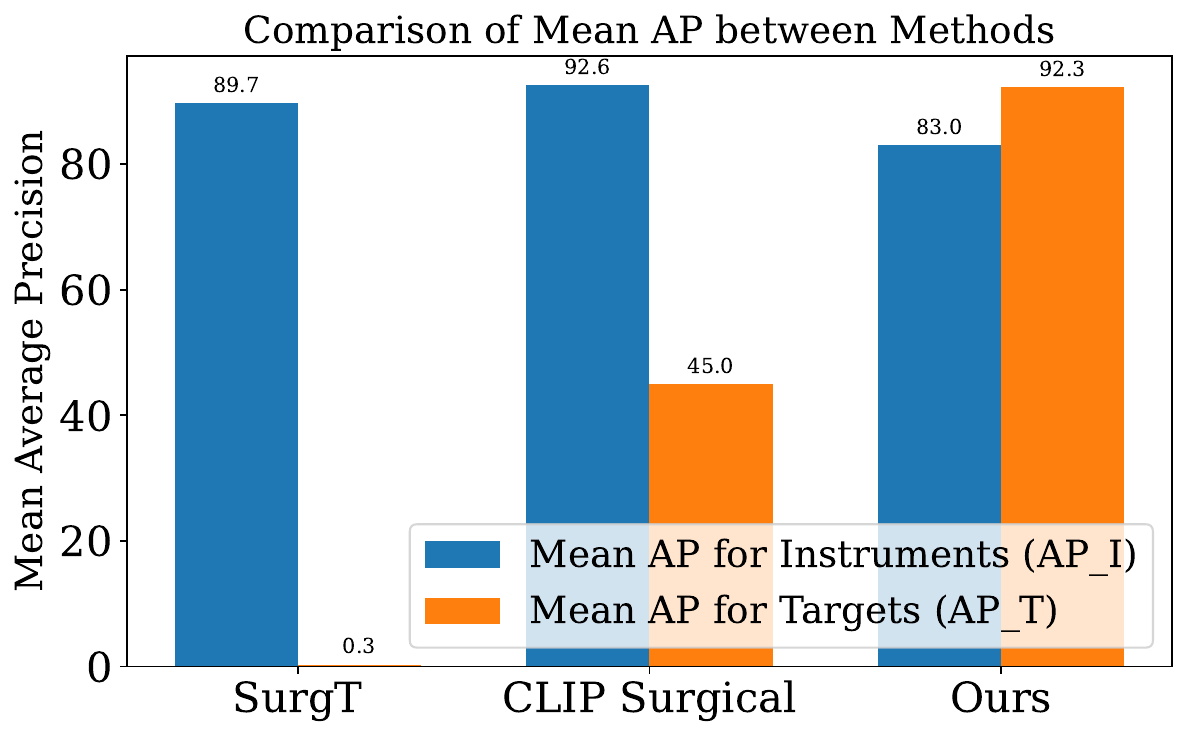} \caption{Mean Average Precision (mAP) comparison across three methods: SurgT, CLIP Surgical, and our proposed approach. Results are reported separately for instrument detection (AP\textsubscript{I}) and target recognition (AP\textsubscript{T}).} \label{fig:mean_AP} \end{figure}

Figure~\ref{fig:mean_AP} presents a comparative bar plot reporting the Mean Average Precision (mAP) for each component of the triplet: instruments (AP\textsubscript{I}) and targets (AP\textsubscript{T}). Our model, labeled as "Ours" in the figure, shows competitive performance overall. In particular, while SurgT and the diffusion-based approach exhibit stronger results for instrument detection, our model significantly outperforms both in the recognition of surgical targets—often the most semantically complex component of the triplet. This indicates a better capability in modeling context-aware information required for target identification.

In summary, the results demonstrate that our model not only provides better-calibrated probability estimates, but also achieves competitive or superior performance in target recognition. This dual benefit—of calibrated confidence and improved semantic understanding—reinforces the suitability of the approach for reliable and interpretable surgical video analysis.

\subsection{Frame Caption Generation}
\label{frame-caption}

To assess the system’s effectiveness in generating automatic frame-level captions, we conducted a comparative evaluation between two configurations of the model: the initial version and a more robust version obtained through fine-tuning. The latter is designed to operate under more realistic conditions, where the input objects are not ground-truth annotations but rather predictions made by an object detector.

The goal of this experiment is twofold: on one hand, to measure the impact of error propagation from the detection module, and on the other, to evaluate the ability of the robust model to adapt to such imperfections while maintaining or improving caption quality. The system's performance is assessed using standard text generation metrics, such as $BLEU$, $ROUGE$, and $BERTScore$.

Table~\ref{tab:frame_results} summarizes the obtained results. As discussed in Section~\ref{sec:training strategy}, the robust model is trained using the predicted objects as input, aiming to learn how to generate accurate captions even in the presence of noisy or partially incorrect input.

The results indicate clear improvements across most metrics. The $BLEU$ score increases from $0.64$ to $0.73$, reflecting better lexical coherence with the reference captions. Similarly, the $ROUGE$ scores show significant gains, indicating improved n-gram coverage of the target phrases. In contrast, $BERT$ Precision remains virtually unchanged, suggesting that even the initial model is capable of producing semantically similar words to the ground-truth descriptions. However, the robust model achieves higher $Recall$ and $F1$ scores, highlighting broader semantic coverage and a better balance between precision and recall.

Although the original model already achieves a solid level of performance, the robust model demonstrates greater resilience to input noise, producing more complete and semantically coherent descriptions. Qualitative examples of captions generated by the robust model are provided in Appendix~\ref{sec:frame captions}, showcasing its ability to capture visual content accurately even under suboptimal conditions.

\begin{table}[!ht]
\begin{adjustwidth}{-0.5in}{-0.5in} 
\centering
\fontsize{8pt}{9pt}\normalfont{
\begin{tabular}{@{\hspace{5pt}}l@{\hspace{5pt}}c@{\hspace{5pt}}c@{\hspace{5pt}}}
\toprule
\multicolumn{1}{c}{\textbf{Metric}} & \textbf{Model} & \textbf{Robust Model} \\
\midrule
BLEU           & 0.6395 & 0.7267 \\
ROUGE-1        & 0.8351 & 0.8700 \\
ROUGE-2        & 0.7747 & 0.8096 \\
ROUGE-L        & 0.8116 & 0.8637 \\
BERT Precision & 0.7771 & 0.7745 \\
BERT Recall    & 0.7644 & 0.8365 \\
BERT F1        & 0.7707 & 0.8052 \\
\bottomrule
\end{tabular}
}
\caption{Frame-level captioning performance of initial vs. robust model using BLEU, ROUGE, and BERTScore.}
\label{tab:frame_results}
\end{adjustwidth}
\end{table}

\subsection{Clip caption Generation}
\label{sec:clip-caption}

Following the evaluation of the system’s ability to generate frame-level captions, a second experiment was conducted focusing on the generation of \textit{clip-level descriptions}. This phase assesses the system's capacity to synthesize extended visual and textual information by summarizing sequences of frames into coherent and semantically meaningful captions. Several models were tested to investigate the impact of different training strategies and input types on the quality of the generated text.

The results, summarized in Table~\ref{tab:clip_results}, are organized based on the type of input used for frame captions: none (unimodal), ground-truth, or automatically generated. Specifically:

\begin{itemize}
    \item The \textbf{Simple} model is a lightweight, unimodal version that only takes the video input, completely ignoring the frame-level captions. This configuration was designed to isolate and evaluate the impact of incorporating intermediate textual information. As expected, it yields the weakest performance across all metrics (e.g., BLEU and BERT Precision), highlighting the limitations of a vision-only approach.
    
    \item The \textbf{Model (GT)} configuration takes ground-truth frame captions as input. This model shows substantial improvements in all evaluated metrics, confirming that high-quality intermediate textual features strongly enhance the performance of the clip-level caption generator.
    
    \item The \textbf{Model (Generated)} variant uses frame captions automatically produced by the frame-level generator. While still performing well, it shows a drop of about 0.02 points in all metrics compared to the ground-truth version. This degradation is likely due to error propagation from the earlier stage. The performance gap is especially notable in semantic metrics such as BERT F1.
    
    \item The \textbf{Robust} model, representing the final and most refined version of the system, achieves the highest scores across nearly all metrics. This model benefits from a two-phase training strategy in which components are initially trained independently and later fine-tuned jointly. This approach allows the system to correct internal inconsistencies and improves robustness to noisy intermediate predictions. It outperforms all other models in BLEU, ROUGE-L, and BERT F1, confirming its capacity to balance lexical accuracy and semantic coherence.
\end{itemize}

\begin{table}[!ht]
\begin{adjustwidth}{-0.5in}{-0.5in} 
\centering
\fontsize{8pt}{9pt}\normalfont{
\begin{tabular}{@{\hspace{5pt}}c@{\hspace{5pt}}c@{\hspace{5pt}}c@{\hspace{5pt}}c@{\hspace{5pt}}c@{\hspace{5pt}}}
\toprule
\multicolumn{1}{c}{\textbf{Frame captions input type}} & \multicolumn{1}{c}{\textbf{None}} & \multicolumn{1}{c}{\textbf{Ground-Truth}} & \multicolumn{2}{c}{\textbf{Generated}} \\
\cmidrule(lr){2-2}\cmidrule(lr){3-3}\cmidrule(lr){4-5}
\textbf{Model type} & \textbf{Simple} & \textbf{Model} & \textbf{Model} & \textbf{Robust} \\
\midrule
BLEU           & 0.5138 & 0.6490 & 0.6023 & 0.6715 \\
ROUGE-1        & 0.7591 & 0.8615 & 0.8317 & 0.8672 \\
ROUGE-2        & 0.6744 & 0.7975 & 0.7447 & 0.7991 \\
ROUGE-L        & 0.7137 & 0.7968 & 0.7605 & 0.8318 \\
BERT Precision & 0.6666 & 0.7733 & 0.6705 & 0.7443 \\
BERT Recall    & 0.6590 & 0.7696 & 0.7486 & 0.7772 \\
BERT F1        & 0.6623 & 0.7714 & 0.7090 & 0.7607 \\
\bottomrule
\end{tabular}
}
\caption{Comparison of model performance depending on training strategy and input type.}
\label{tab:clip_results}
\end{adjustwidth}
\end{table}

These findings emphasize the importance of progressively integrating the system's components and training with realistic, internally generated inputs. The robust model, through its hierarchical and error-aware training process, produces more accurate, fluent, and resilient descriptions—making it particularly well-suited for surgical video summarization.

\subsection{Evaluation of Structured Surgical Report Generation}
\label{sec:ev-report-generation}

To evaluate the final stage of the pipeline—that is, the generation of a structured surgical report from clip-level descriptions—we conducted a qualitative analysis of the produced narrative. The report was synthesized using GPT-4, as described in Section~\ref{sec:Pre-trained Models}, through a carefully designed prompt aimed at merging sequential descriptions into a clinically coherent summary.

The main objectives of this evaluation were to verify that the generated report maintains temporal consistency with the original surgical procedure, while also ensuring that clinically relevant details—such as tool usage, anatomical references, and transitions between surgical phases—are clearly and accurately included. Additionally, the evaluation aimed to assess the overall fluency, readability, and alignment of the report with the standard conventions of surgical documentation.

The results show that the GPT-based summarization module effectively consolidates and structures the information encoded in the clip-level descriptions. The model is able to aggregate durations of clips that describe the same activity and to produce clear transitions between the different phases of the procedure. Importantly, despite some lexical or syntactic imperfections in the intermediate captions (see Appendix~\ref{sec:frame captions}), the final report mitigates these issues by leveraging the language model’s contextual reasoning capabilities and its ability to eliminate redundancy.

A complete example of a report generated by our system is provided in Appendix~\ref{sec:report-example}, corresponding to video \texttt{VID07}. This example illustrates how our pipeline transforms segmented and multimodal inputs into a coherent surgical narrative, thereby demonstrating the feasibility of using LLMs for post-hoc summarization in high-stakes clinical settings.

\section{Conclusion}
\label{sec:conclusion}

The integration of artificial intelligence into surgery has the potential to enhance precision, decision-making, and documentation. However, current AI models, particularly convolutional neural networks (CNNs), often struggle with capturing long-range dependencies, a crucial aspect of understanding complex surgical scenes. In this work, we address this challenge by combining recent advancements in computer vision and natural language processing to automatically generate detailed surgical reports from video data. Our approach leverages the strengths of three specialized components—object detection, visual feature extraction, and temporal modeling—within a modular architecture. This design facilitates comprehensive scene understanding while limiting visual hallucinations and enabling more interpretable outputs.

Experimental results demonstrate the effectiveness of our methodology. The incorporation of multi-turn reasoning was shown to improve the contextual understanding of surgical videos, leading to performance gains of up to 12\% in semantic precision and 30\% in BLEU score. These improvements highlight the model's capacity to generate accurate and coherent textual summaries by breaking down complex scenes into manageable subtasks, each handled by a dedicated module.

Despite these encouraging results, several avenues remain for further enhancement. Expanding the dataset to encompass a wider variety of surgical procedures would significantly improve the model's generalization capability and lay the groundwork for developing a robust, domain-specific foundation model. Increasing the temporal range analyzed by the vision module could allow for better modeling of long-range dependencies, thereby improving the understanding of complex procedural dynamics. Moreover, incorporating additional modalities—such as object segmentation for spatial awareness or audio cues from surgical tools for action recognition—could enrich the contextual representation and lead to more precise and informative report generation.

Although the current pipeline does not yet rely on autonomous agents, its modular and task-specific design is well-suited for future integration into agent-based architectures. In such frameworks, autonomous AI agents could execute entire surgical video analysis workflows independently—coordinating detection, captioning, and summarization tasks—while continuously optimizing the process based on context and feedback. When combined with explainable AI (XAI) techniques, these systems would not only provide transparency into each step of the reasoning process but also empower users to oversee and influence strategic decision-making. This could include setting operational boundaries, controlling the behavior of individual agents, and maintaining a human-in-the-loop paradigm, which is particularly critical in high-stakes domains like surgery where trust, safety, and accountability are paramount.

\bibliographystyle{plainnat}
\bibliography{references}
\clearpage

\appendix

\begin{appendices}
\section{Frame Captions}
\label{sec:frame captions}

In this section are presented some generated caption examples. Despite the positive results presented in section \ref{sec:experiments}, the model has trouble generating the word $gallbladder$ and sometimes generates $"gallbloddger"$ instead as in the second example. This issue is likely due to tokenization, where the model may incorrectly split or merge tokens, leading to these small mistakes. \\
\rule{\linewidth}{0.1mm} \\
\textbf{Predicted}: During phase carlot-triangle-dissection, the grasper is grasping the gallbladder, their bipolar is dissecting the cystic\_artery \\
\textbf{Target}:  During phase carlot-triangle-dissection, the bipolar is dissecting the cystic\_artery, the grasper is grasping the gallbladder \\
\rule{\linewidth}{0.1mm}  \\
\textbf{Predicted}: During phase gallbladder-dissection, the hook is dissecting the gallbloddger \\
\textbf{Target}:  During phase gallbladder-dissection, the hook is dissecting the gallbladder \\
\rule{\linewidth}{0.1mm}  \\
\textbf{Predicted}: During phase carlot-triangle-dissection, the grasper is retracting the gallbladder, this hook is present \\
\textbf{Target}:  During phase carlot-triangle-dissection, the grasper is retracting the gallbladder, the hook is present\\
\rule{\linewidth}{0.1mm}  \\
\textbf{Predicted}: During phase carlot-triangle-dissection, the grasper is retracting the gallbladder, those bipolar is coagulating the abdominal\_wall\_cavity \\
\textbf{Target}:  During phase carlot-triangle-dissection, the grasper is retracting the gallbladder, the bipolar is coagulating the abdominal\_wall\_cavity \\
\rule{\linewidth}{0.1mm}  \\
\textbf{Predicted}: During phase carlot-triangle-dissection, the grasper is retracting the gallbladder, this hook is dissecting the omentum \\
\textbf{Target}:  During phase carlot-triangle-dissection, the grasper is retracting the gallbladder, the hook is dissecting the omentum

\section{Report Example}
\label{sec:report-example}
Report example for the video $VID07$ using $GPT-4$ from open AI with the clip descriptions and the prompt as input. \\ 

\texttt{ Report for VID07:\\  \rule{\linewidth}{0.1mm}   \textcolor{white}{a} \\
The surgical procedure initiated with the preparation phase, which lasted for a total of 56 seconds. Initially, the grasper was engaged in retracting the gallbladder while the hook was used to dissect the gallbladder. Following this, the clipper retracted the omentum for 8 seconds. \\
The next phase was the Calot triangle dissection, which took an extended duration due to the complexity involved. The hook was used to dissect the gallbladder while the grasper helped in retracting the gallbladder. The dissection of cystic artery and cystic duct was carried out multiple times during this phase. The grasper maintained a consistent presence, assisting in retracting the gallbladder throughout.\\
The procedure moved on to the clipping and cutting phase which lasted for 117 seconds in total. The grasper was utilised to retract the gallbladder while the clipper was engaged in clipping the cystic artery and the cystic duct. The scissors were also used to cut the cystic duct and artery.\\
Following this, the gallbladder dissection phase was carried out for a prolonged duration. The hook dissected the gallbladder, and the grasper retracted the gallbladder and the liver simultaneously. During this phase, the irrigator was also used to retract the liver while the grasper was present.\\
The gallbladder packaging phase took 48 seconds. The grasper was engaged in packing the gallbladder, and the liver was retracted. The specimen bag was handled by the grasper during this phase.\\
The cleaning and coagulation phase took place next. The grasper was used to retract the liver, the irrigator was employed to aspirate the fluid, and the bipolar was used to coagulate the omentum. There were instances where the omentum was aspirated and the abdominal wall cavity was irrigated.\\
The final phase was the gallbladder extraction, which lasted for 79 seconds. The grasper was present throughout, the hook was used for gallbladder dissection, and the scissors were used for handling the specimen bag. The procedure concluded with the extraction of the gallbladder.
}

\section{Models Parameters}
\label{sec:modelparameters}

\textcolor{white}{a} \\
\texttt{--------------------------------------------------------------------------------------------------------------------------------\\
Object Detector\\ --------------------------------------------------------------------------------------------------------------------------------\\
        Layer (type) \ \ \ \ \ \ \ \ \ \ \ \ \ \ \ \ Output Shape \ \ \ \ \ \ \ \  \ \ \ \ \ \ \ Param \#   \\
================================================================   \\
   ViTModel-1\ \ \ \ \ \  \ \ \ \ \ \ \ \ \ \ \ \ \ [-1, 196, 768] \ \ \ \ \ \ \ \ \ \ \ \ 85207296      \\
   Linear-2\ \ \ \ \ \ \ \ \ \ \ \ \ \ \ \ \ \ \ \ \ [-1, 21]      \ \ \ \ \ \ \ \ \ \ \ \ \ \ \ \ \ \ \ \ \  16149       \\
================================================================\\
Total params: 85223445\\
Trainable params: 85223445\\
Non-trainable params: 0\\
--------------------------------------------------------------------------------------------------------------------------------
}
\textcolor{white}{a} \\
\texttt{--------------------------------------------------------------------------------------------------------------------------------\\
Frame Captioner\\ --------------------------------------------------------------------------------------------------------------------------------\\
        Layer (type) \ \ \ \ \ \ \ \ \ \ \ \ \ \ \ \ Output Shape \ \ \ \ \ \ \ \  \ \ \ \ \ \ \ Param \#   \\
================================================================   \\
   ViTModel-1\ \ \ \ \ \  \ \ \ \ \ \ \ \ \ \ \ \ \ [-1, 196, 768] \ \ \ \ \ \ \ \ \ \ \ \ 86389248      \\
   Linear-2\ \ \ \ \ \ \ \ \ \ \ \ \ \ \ \ \ \ \ \ \ [-1, 196, 512]      \ \ \ \ \ \ \ \ \ \ \ \ \ \  393728     \\
   DistilBertModel-3 \ \ \ \ \ \ \ \ \ \ \  [-1, 64, 768]  \ \ \ \ \ \ \ \ \  \ \ \ \ 66362880 \\
   Linear-4\ \ \ \ \ \ \ \ \ \ \ \ \ \ \ \ \ \ \ \ \ [-1, 64, 512]          \ \ \ \ \ \ \ \ \ \ \ \ \ \ \ 393728\\
   Fusion-5\ \ \ \ \ \ \ \ \ \ \ \ \ \ \ \ \ \ \ \ \ [-1, 260, 512]          \ \ \ \ \ \ \ \ \ \ \ \ \ \ \ \ \ \ \  0\\
   T5ForConditionalGeneration-6  [-1, num\_tokens, 32128]\ \ \ \ \ 60506624  \\
================================================================\\
Total params: 214046208\\
Trainable params: 214046208\\
Non-trainable params: 0\\
--------------------------------------------------------------------------------------------------------------------------------
}
\textcolor{white}{a} \\
\texttt{--------------------------------------------------------------------------------------------------------------------------------\\
Clip Captioner\\ --------------------------------------------------------------------------------------------------------------------------------\\
        Layer (type) \ \ \ \ \ \ \ \ \ \ \ \ \ \ \ \ Output Shape \ \ \ \ \ \ \ \  \ \ \ \ \ \ \ Param \#   \\
================================================================   \\
   VivitModel-1 \ \ \ \ \ \ \ \ \ \ \  \ \ \ \ \ [-1, 3137, 768] \ \ \ \ \ \ \ \ \ \ \ 89236992      \\
   Linear-2\ \ \ \ \ \ \ \ \ \ \ \ \ \ \ \ \ \ \ \ \ [-1, 3137, 768]       \ \ \ \ \ \ \ \ \ \ \ \ \  590592   \\
   DistilBertModel-3 \ \ \ \ \ \ \ \ \ \ \  [-1, 4096, 768] \ \ \ \ \ \ \ \ \  \ \ 66362880 \\
   Linear-4\ \ \ \ \ \ \ \ \ \ \ \ \ \ \ \ \ \ \ \ \ [-1, 4096, 768]          \ \ \ \ \ \ \ \ \ \ \ \ \ 590592\\
   Fusion-5\ \ \ \ \ \ \ \ \ \ \ \ \ \ \ \ \ \ \ \ \ [-1, 7233, 768]          \ \ \ \ \ \ \ \ \ \ \ \ \ \ \ \ \ \ 0\\
   T5ForConditionalGeneration-6 [-1, num\_tokens, 32128]\ \ \ \ 248168448 \\
================================================================\\
Total params: 404358912\\
Trainable params: 404358912\\
Non-trainable params: 0\\
--------------------------------------------------------------------------------------------------------------------------------
}

\end{appendices}

\end{document}